\documentclass{article}

\pdfoutput=1
\usepackage{arxiv}

\usepackage[utf8]{inputenc} 
\usepackage[T1]{fontenc}    
\usepackage{hyperref}       
\usepackage{url}            
\usepackage{booktabs}       
\usepackage{amsfonts}       
\usepackage{nicefrac}       
\usepackage{microtype}      

\usepackage{graphicx}
\usepackage{amsmath,amssymb} 
\usepackage{color}

\usepackage{subcaption}
\usepackage{adjustbox}

\usepackage{tablefootnote}

\title{Visual recognition in the wild by sampling deep similarity functions}

\author{
  Mikhail Usvyatsov\\
   Department of Civil, Environmental and Geomatic Engineering\\
  ETH Zurich, Switzerland \\
  \texttt{mikhail.usvyatsov@geod.baug.ethz.ch} \\
   \And
 Konrad Schindler \\
 Department of Civil, Environmental and Geomatic Engineering\\
ETH Zurich, Switzerland \\
  \texttt{konrad.schindler@geod.baug.ethz.ch} \\
}

\begin{document}
\maketitle

\begin{abstract}
 Recognising relevant objects or object states in its environment is
a basic capability for an autonomous robot. The dominant approach to
object recognition in images and range images is classification by
supervised machine learning, nowadays mostly with deep convolutional
neural networks (CNNs). This works well for target classes whose
variability can be completely covered with training
examples. However, a robot moving in the wild, i.e., in an
environment that is not known at the time the recognition system is
trained, will often face \emph{domain shift}: the training data
cannot be assumed to exhaustively cover all the within-class
variability that will be encountered in the test data. In that
situation, learning is in principle possible, since the training set
does capture the defining properties, respectively dissimilarities,
of the target classes. But directly training a CNN to predict class
probabilities is prone to overfitting to irrelevant correlations
between the class labels and the specific subset of the target class
that is represented in the training set. We explore the idea to
instead learn a Siamese CNN that acts as similarity function between
pairs of training examples. Class predictions are then obtained by
measuring the similarities between a new test instance and the
training samples. We show that the CNN embedding correctly recovers
the relative similarities to arbitrary class exemplars in the
training set. And that therefore few, randomly picked training
exemplars are sufficient to achieve good predictions, making the
procedure efficient.
\end{abstract}

\section{Introduction}
Deep learning, and in particular convolutional neural networks, have
revolutionised the field of visual recognition. However, the
break-through has been achieved under the assumption of a stable domain,
meaning that we have access to a sufficiently large training database
that covers \emph{all} the modes of variation of the target classes
that we might encounter at test time.
In practice, this is achieved either by operating in a constrained
environment in which we can indeed exhaustively sample the expected
visual variability; or by collecting and labeling very large datasets,
in an attempt to brute-force the generalisation problem via a training
set that exhaustively exposes the variability of the entire ``visual
world''.
This procedure has enjoyed immense success for tasks where the class
appearance is fairly static, and where finding data and labeling them
is easy, if tedious (such as recognising cats in internet images, or
cars in images taken by an autonomous vehicle). However, there are
problems where it is not as easy to obtain ground truth that
generalises to all relevant cases. For instance, some variations in
the target class may be rare or access to them may be restricted. More
importantly, there are situations where the target class undergoes
\emph{domain shift}, such that it is not only difficult, but
impossible to know all admissible modes of variation in advance, and
at test time we are bound to encounter examples which are not well
represented in the training distribution.
As a concrete example, imagine a robotic watchman that shall be
shipped with the capability to recognise whether a door is open or
closed (Fig.~\ref{fig:0}), in an a-priori unknown environment. The
road-block for training a classifier is not that we cannot find enough
images of doors. Rather, it is that new doors with different
appearance are designed and built all the time.
%
%
\makeatletter
\newlength{\robotwidth}
\setlength\robotwidth{0.9\columnwidth}
\makeatother
\begin{figure}[t]
	\centering
	\includegraphics[width=\robotwidth]{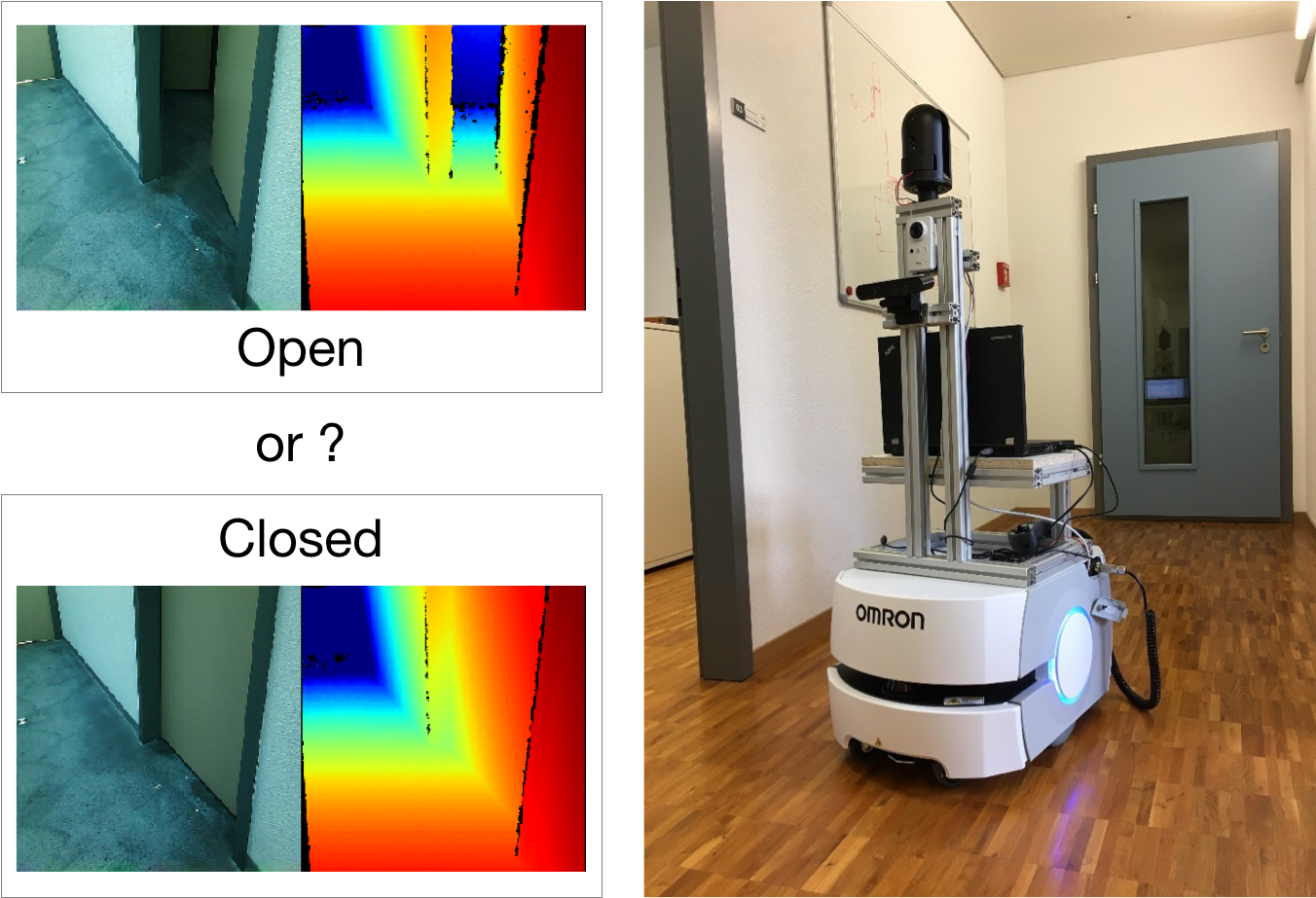}
	\caption{The setting: a robot shall classify objects -- here
		open vs.\ closed doors -- in a new environment, where their
		visual appearance may differ from the training set.}
	\label{fig:0}
	\vspace{-0.5em}
\end{figure}
The problem is related to few-shot learning, although we are in the
domain-shift scenario not adding a new, previously unseen class; in
the sense that the relevant information to learn a suitable
representation for the classification task is actually present in the
data, but the learning is unable to separate important discriminative
properties from irrelevant intra-class variation.
Few-shot learning, say adding a new breed of dog to a classifier, aims
to learn an embedding of the input images in which different classes
form clusters, such that inter-class distances are higher than
intra-class distances. The hope is that embedding a new class will
lead to a new cluster, because the embedding generalises across
different dog breeds, whereas a purely discriminative scheme could not
be trained from only few examples. Classification with the learned
distance function is achieved by computing distances between training
and test examples, followed by nearest-neighbour classification.

We face a similar situation: pictures of open doors do have
characteristic properties (like the bottom edge of the door not
connecting the bottom ends of the frame). The problem is not that the
crucial discriminative properties cannot be learned -- they \emph{are}
captured in the training set. Rather, a straight-forward two-class CNN
does not learn well enough to ignore legitimate intra-class variations
not present in the training examples, i.e., it overfits to the
training domain. Pre-training on a larger dataset and fine-tuning to
the task-specific training dataset will possibly mitigate, but not
solve the problem: there is no reason why pre-training on, say,
ImageNet would inject exactly the information that is needed to
generalise to the unseen doors (which are not observed in ImageNet,
either). Likewise, heavier regularisation can potentially mitigate
overfitting to biases in the training set, but empirically does not
solve the problem -- which is not that surprising: the regulariser
favours similar outputs for nearby data points, but the domain shift
is caused by examples that are far from all training samples, and thus
calls for a notion of similarity that is valid everywhere on the input
manifold.

We propose an approach based on a Siamese network for similarity
learning. Our method is technically related to recent works on
one-shot or few-shot deep
learning~\cite{koch2015siamese,bertinetto2016learning,vinyals2016matching,sung2018learning},
but differs from them in one important aspect: few-shot learning aims
to learn a \emph{new} class from only few training examples (often
motivated by the astonishing capabilities of biological vision),
whereas we are concerned with generalisation of the \emph{same} class
beyond the domain covered by a (possibly not so small) training
set. Computing similarities to all training examples for each test
sample is computationally expensive. We show that it can be replaced
by a $\mathcal{O}(1)$ random sampling procedure without performance
penalty.
In experiments on a diverse collection of datasets, our
method consistently outperforms the direct classification baseline,
reducing classification errors by 28-64\%.

\section{Related Work}

\noindent
\textbf{Distance learning.}
In their comprehensive survey \cite{yang2006distance}, Yang et al.\
classify distance learning methods into three groups,
\emph{(i)}~unsupervised distance learning, \emph{(ii)}~supervised
global distance learning, and \emph{(iii)}~supervised local distance
learning.
The idea of unsupervised methods is to learn a lower-dimensional
embedding that preserves the pairwise distances between data
points. Besides classical techniques like PCA and its non-linear
generalisations (KPCA~\cite{schoelkopf1998}, LLE~\cite{roweis2000},
etc.), unsupervised distance learning can also be implemented with
neural networks, e.g., auto-encoders \cite{maaten2009learning}.

Supervised methods are discriminative in that they use class
labels to build equivalence constraints between objects in a
dataset.  Global methods try to enforce all the constraints
simultaneously, aiming for a globally valid embedding where
within-class distances are smaller than between-class
distances. There are several ways to turn this requirement into a
loss function and minimise it, e.g.,
\cite{xing2003distance,lebanon2003flexible}.

Local methods take a more flexible approach and only demand that a
local subset of the equivalence constraints are fulfilled, so as to
enable classification based on local neighbours \cite{yang2006distance}.
In that view, Siamese network approaches could be called local
methods, since only a small set of distances to the few available
class exemplars must be correctly preserved to achieve, for instance,
few-shot classification via $k$NN. Whereas distances to other class
members not used as exemplars could in principle be arbitrarily
distorted without influencing the result.

Our work, however, suggests that a Siamese network in fact achieves an
\emph{approximately global} embedding: While it is still unclear how
well the global distance ordering is preserved, our experiments show
that the distances to members of the correct class are on average
lower than to members of the incorrect one, for arbitrary subsets of
training examples.
This implies that the distance computation does not depend on using
fixed reference exemplars for a class, and that it can be robustified
by sampling and averaging.
Our way of using the embedding provided by the Siamese network is thus
orthogonal to few-shot learning with a fixed, small set $k$ of
exemplars per class. It indicates that learning the embedding rather
than the labeling can also be beneficial when there are more than a
few examples; and that one actually need not commit to a fixed set of
exemplars at all, if more data is available.


\vspace{0.5em}
\noindent
\textbf{Siamese networks for distance learning.}
A Siamese network is a neural network that has multiple inputs of the
same size, which are processed by identical branches with shared
weights before combining them to generate the desired output.
Bromley et al.\ first introduced the notion of Siamese networks, in
the context of signature verification
\cite{bromley1994signature}. They proposed to learn a pairwise
distance measure between feature vectors (in their case derived from
time-series of $x,y$-coordinates on a tablet) that represent two
different objects (in their case signatures).

After the advent of modern convolutional networks, the same idea was
applied to raw images,
e.g.,~\cite{zbontar2015computing,lin2015learning,zagoruyko2015learning,luo2016efficient,hartmann2017learned}. Siamese
convolutional branches independently transform two (or more) images
$A$ and $B$ into high-level representations that are then merged and
transformed further into a learned measure $F(A,B)$ of similarity.%
\footnote{Note, $F$ is normally not a metric in the formal sense.} %
Koch et al.~\cite{koch2015siamese} were perhaps the first to use
Siamese networks for one-shot learning, using the learned
image-to-image similarity in conjunction with an exemplar for each
of the target classes to perform nearest-neighbour classification.
Similar ideas are also elaborated in
\cite{vinyals2016matching,sung2018learning}. The approach naturally
covers also few-shot learning, by using consensus voting over the
similarities to $k$ exemplars per class.
However, the number $k$ must remain small, otherwise the method
quickly becomes inefficient, because the similarity computation for
each individual exemplar amounts to a complete forward pass of the
network (e.g., with our architecture based on VGG16, $\approx
0.015\cdot k$ seconds on an Nvidia Titan Xp GPU).

\section{Method}

\noindent
\textbf{Deep similarity learning.}
The baseline for supervised classification is to directly predict the
class label for an input image. In the following, we limit the
discussion to binary classification, the extension to multiple classes
is straight-forward. Hence, we start from images
$X\in\mathbb{R}^{N\times M\times D}$ and corresponding labels
$y\in\{0,1\}$ and fit a mapping
\begin{equation}
y=f(X)\quad,\quad y\in[0\hdots 1]\quad,
\label{eq1}
\end{equation}
where $y$ is a soft score between $0$ and $1$ that can be interpreted
as the probability that the image belongs to class~$1$. A decision
rule is obtained by simply thresholding $y$. In image classification,
the state of the art for the function $f$ are deep convolutional
neural networks (CNNs).

That baseline is purely discriminative and has no notion of
distance in the input space. It performs exceedingly well in stable
domains (which includes typical benchmark datasets). But in the
presence of domain shift its unrivalled ability to discover any
discriminative pattern in the input becomes a liability: as soon as
the available training set exhibits not only the relevant patterns
(e.g., the characteristic differences between open and closed doors)
but also other, spurious ones with even weak predictive power, the
network is going to detect those and overfit to them.

Instead, we propose to resort to \emph{similarity learning}.%
\footnote{We avoid the term ``distance learning'',
	since outputs are $\tilde{y}\in[0\hdots 1]$.}
Here, the input are pairs of
images $\tilde{X}\in\mathbb{R}^{N \times M \times D} \times
\mathbb{R}^{N \times M \times D}$, taken either from the same class
($\tilde{y}=1$) or from different classes ($\tilde{y}=0$). To these,
we fit a function
\begin{equation}
\tilde{y}=F(\tilde{X})\quad,\quad \tilde{y}\in[0\hdots1]\quad.
\end{equation}
This is still a binary classification problem and can be trained with
the same loss function as direct classification. But the output has a
different meaning: previously we predicted the probability of one
input image to belong to class $1$, thus potentially including
spurious correlations due to unintended biases in the training
data. While in the new formulation we predict the probability that the
two images of a new pair are of the same class, thus focussing on
whether the critical markers for being different are present or
absent.
$\tilde{y}=1$ denotes maximal similarity, $\tilde{y}=0$
maximal dissimilarity.

\begin{figure}
	\includegraphics[width=\linewidth]{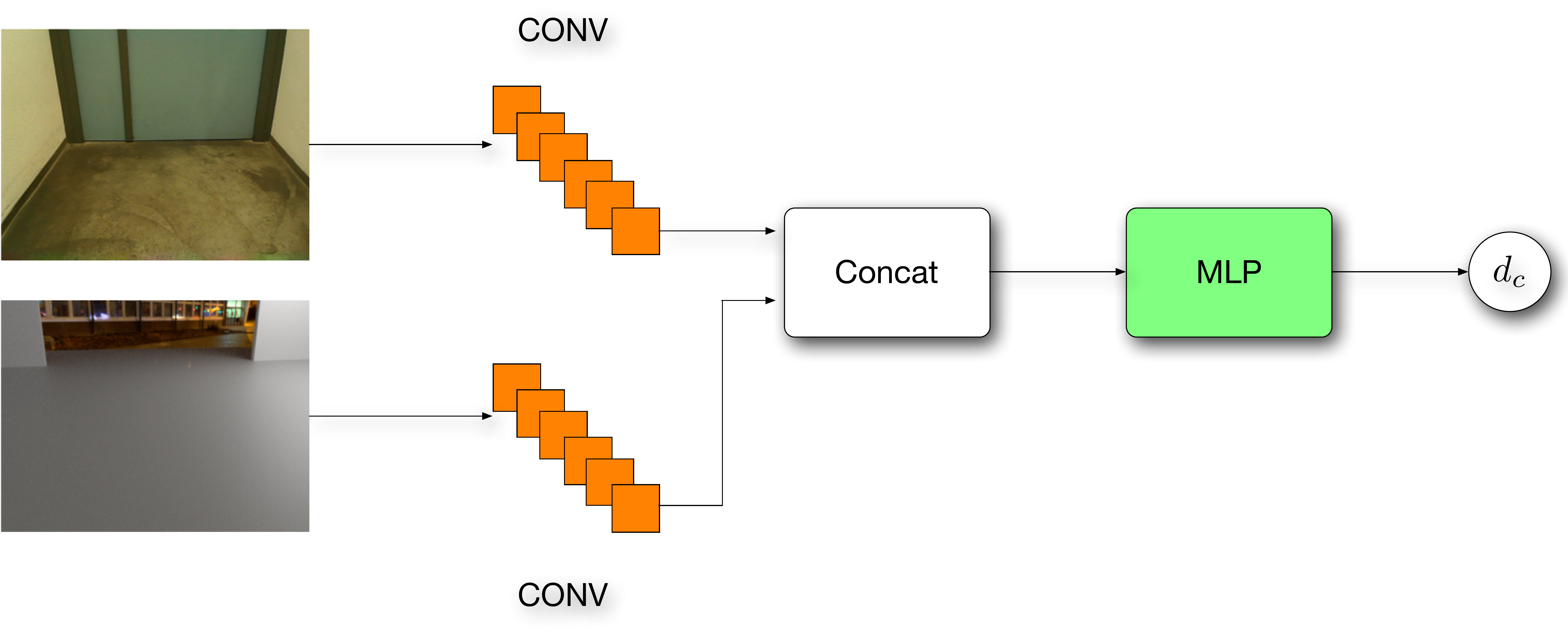}
	\caption{Siamese network: given two input images (top: closed
		door, bottom: open door), extract features from both images
		using convolutional branches with tied weights, followed by
		a classification stage that predicts the probability that
		the inputs are from different
		classes.}
	\label{fig:1}
	\vspace{-0.5em}
\end{figure}

At first glance, it may seem that by moving from single images to
image pairs as input, we are solving a more difficult learning
problem.
But in fact, the Siamese network is not more complex. Its number of
parameters is similar to the conventional classification network, it
outputs only a scalar value, and by looking at pairs we have
quadratically increased the number of training samples. Importantly,
sampling pairs of training images reduces spurious correlations. As a
simplified cartoon example, consider a case where one can see the sky
behind the door in many training examples of the ``open'' state. A
direct classifier will learn that a blue region boosts the score
for ``open'' -- which is correct for the training set, but may hurt
performance in a previously unseen environment with blue
doors. Whereas the pair classifier can at most learn that a blue
region being present in only one of the two images boosts the score
for ``different'', which is likely to be true even in the new
environment.

\vspace{0.5em}
\noindent
\textbf{Efficient similarity-based classification.}
An important question is how to utilise the learned similarity
function in an efficient manner to classify an unseen test example
$Q$. The obvious approach would be some sort of $k$-nearest neighbour
scheme.  However, that is fairly inefficient, because to find the
nearest neighbours one has to compute the similarities from the test
sample to \emph{all} training examples, each corresponding to a
forward pass of the network. Alternatively, one can chose one (or a
few) representatives for each class and compute the similarities only
to them, as often done in few-shot classification. By representing
each class with a single representative, one potentially sacrifices
robustness to gain speed. Particularly when working with highly
non-linear deep embeddings, it is not obvious how to find the best
representatives. In fact it is not even clear that any fixed, small set
of exemplars works for all test examples.

A main finding of the present work is that it is not necessary to find
a privileged set of ``nearest'' or ``suitable'' class exemplars for
similarity computation.
Rather, we observe that good results are achieved by using the average
similarity to any random subset of class representatives in the training
set. This includes the (inefficient) extreme case of using the average
similarity over all class exemplars, as well as the opposite extreme of
blindly sampling a single exemplar per class for every new query.
It thus appears that the CNN embedding preserves, at least
approximately, a \emph{global} notion of distance; and that it indeed
manages to separate the classes by wide margins, such that most
individual similarities yield a correct class assignment.

\makeatletter
\newlength{\charwidth}
\setlength\charwidth{0.2\columnwidth}
\setlength\tabcolsep{0pt}
\makeatother
\begin{figure}[t]
	\centering
	\begin{tabular}{ccccc}
		\toprule
		{\small exemplars} & {\small similarity} & {\small test example} & {\small similarity} & {\small exemplars}\\
		{\small Katakan}a & $\tilde{y}$ & {\small (Katakana)} & $\tilde{y}$ & {\small Korean}\\
		\midrule
		\adjustbox{valign=m}{\includegraphics[width=\charwidth]{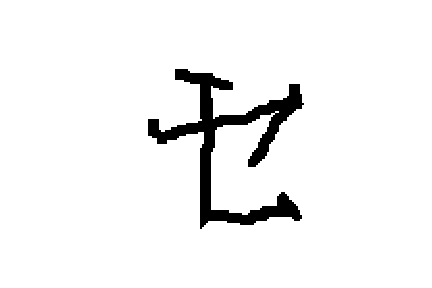}} &
		$\begin{matrix} 0.7 \\ \longleftrightarrow \end{matrix}$ & &
		$\begin{matrix} 0.1  \\ \longleftrightarrow \end{matrix}$ &
		\adjustbox{valign=m}{\includegraphics[width=\charwidth]{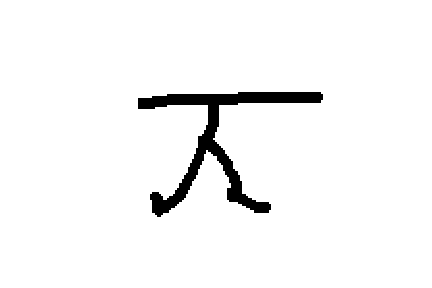}}\\
		\adjustbox{valign=m}{\includegraphics[width=\charwidth]{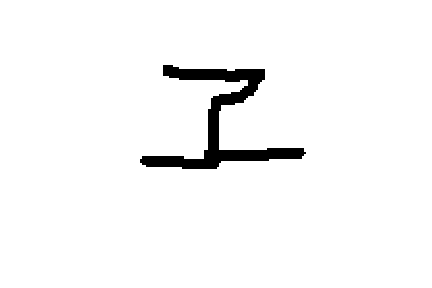}} &
		$\begin{matrix} 0.8 \\ \longleftrightarrow \end{matrix}$ &
		\adjustbox{valign=m}{\includegraphics[width=\charwidth]{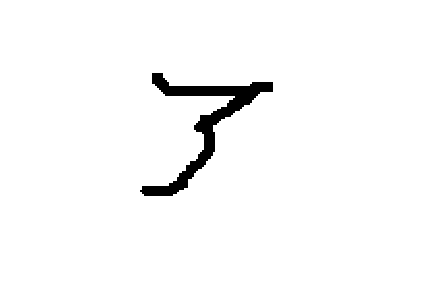}} &
		$\begin{matrix} 0.7  \\ \longleftrightarrow \end{matrix}$ &
		\adjustbox{valign=m}{\includegraphics[width=\charwidth]{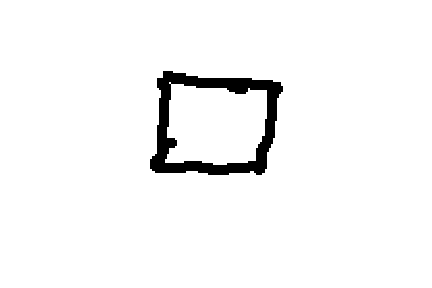}}\\
		\adjustbox{valign=m}{\includegraphics[width=\charwidth]{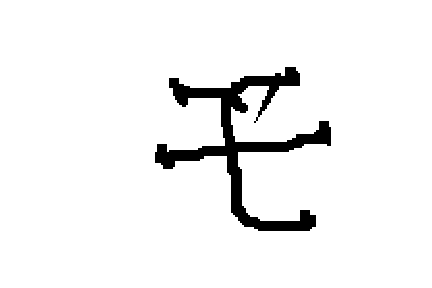}} &
		$\begin{matrix} 0.1 \\ \longleftrightarrow \end{matrix}$ & &
		$\begin{matrix} 0.4  \\ \longleftrightarrow \end{matrix}$ &
		\adjustbox{valign=m}{\includegraphics[width=\charwidth]{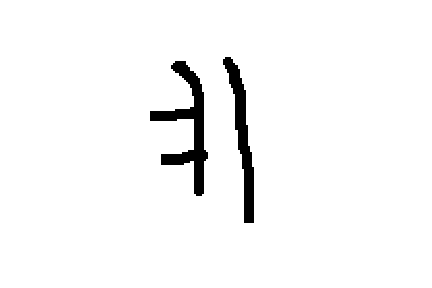}}\\
		\midrule
		&
		$S_{kata}=0.53$ & &
		$S_{kor}=0.4$ & \\
		\bottomrule
	\end{tabular}
	\caption{Classification by averaging similarities to random
		exemplars. In the example, an unseen character shall be classified
		as belonging to the \emph{Katakana} or the \emph{Korean}
		alphabet.}
	\label{fig:4}
	\vspace{-0.5em}
\end{figure}

Technically, one simply samples $k$ representatives $\{R^c_j,j=1\hdots
k\}$ at random from the training samples $\{X^c_i\}$ of each class
$c$.
Similarities $\tilde{y}^c_j=F(Q,R^c_j)$ are computed for all of them
by passing $Q$ and $R^c_j$ through the Siamese network. This yields
$k$ similarity scores per class, which are averaged to obtain a
classification score:
\vspace{-0.5em}
\begin{equation}
S_c(Q)=\frac{1}{k}\sum_{j=1}^k F(Q, R^c_j)
\label{eq6}
\vspace{-0.5em}
\end{equation}
The classification rule then assigns the class with the higher average
score, $\text{argmin}\big(S_0(Q),S_1(Q)\big)$. The procedure is
illustrated in Fig.~\ref{fig:4}.

Obviously, one could think of many other possible consensus
mechanisms. E.g., one might interpret the CNN output $\tilde{y}$ as
probability of being in the same, respectively different classes,
threshold each of the $k$ individual scores with a threshold of $0.5$,
and then perform majority voting (this is sometimes done in ensemble
classifiers, for instance Random Forests).
We experimented also with the latter scheme and found that it performs
similar to averaging, see Table~\ref{table1}. Actually, the fact that
even sampling a single random exemplar per class ($k=1$) works well
indicates that the embedding separates the classes with a healthy
margin, such that voting will also work.
We note that voting may be less robust in some situations. The early
rounding from soft to hard, binary similarity scores may be
problematic if the scores are not well calibrated to probabilities,
such that $0.5$ is not the optimal threshold. Moreover, voting with
few exemplars can lead to ties.
Further research is required, in particular one might also try to
learn the consensus mechanism along with the Siamese embedding.
At this point, we prefer averaging, which appears to be the safer
option when applied to an unknown dataset.


\vspace{0.5em}
\noindent
\textbf{Architectures.}
Both our baseline and our Siamese similarity network are based on the
VGG16 architecture. The Siamese variant has two tied VGG16
branches. Their outputs are concatenated (subtracting them works
equally well) and fed through a multi-layer perceptron with three
fully connected layers to obtain the final scores.
Training is done with the ADAM variant of stochastic gradient descent,
with minibatches of size 16 for the Siamese network, respectively 32
for the single-branch baseline.
The smaller batch size is meant to ensure a fair comparison in terms
of ressources, since image pairs need twice as much memory. GPU memory
is the bottleneck for CNN training when working with large images
(like our ``doors'' dataset).
For the ``Learning to Compare'' baseline we use the architecture and
hyper-parameters of the original, publicly available code.

\vspace{0.5em}
\noindent
\textbf{Pre-training.}
It is standard practice to pre-train deep neural networks with very
large databases to improve their performance, even if those databases
are not perfectly matched to the actual task.
The pre-training is completed by the actual training with a smaller
amount of task-specific data.

Also for the Siamese network, existing, large databases can be
exploited for pre-training.
Empirically, pre-training the individual branches with a conventional
classification task (respectively, using pre-trained layers from
standard classification architectures) does not improve over random
initialisation.
On the contrary, it is beneficial to pre-train the similarity network
with external data, unless one has a very large training set. For our
purposes, we used ImageNet to randomly generate pairs belonging to the
same class, respectively different classes. We pre-train the network
with these pairs to predict the probability that the two inputs belong
to different classes, using the conventional cross-entropy loss.
Then, the same pairwise similarity training is repeated with the actual
training data of the target problem to fine-tune to the application
setting.
In our experiments, pre-training improves performance on all
datasets, by $\approx$10 percent points. For completeness, we
note that \emph{only} ImageNet pre-training, without subsequent tuning
on task-specific data, is not enough and barely better than random
chance.

\section{Experiments}

We conducted experiments on four different datasets (greyscale,
RGB, and RGB-D) to evaluate the proposed classification scheme.
The consistent improvements on a wide variety of datasets and inputs
indicates that the method is fairly general and not limited to
particular classes or image characteristics.
For each dataset we construct a binary task that corresponds to our
goal of operating under domain-shift, i.e., the training set shows
how to separate the classes, but does not show the full
within-class variability.
Beyond our actual application of doors, we used public datasets to
ensure the results are repeatable and can be compared against. Those
public datasets were not designed with domain shift in mind. We did
our best to design tasks that are challenging and representative of
real applications.

\makeatletter
\setlength\tabcolsep{2pt}
\makeatother
\begin{table}
	\centering
	\begin{tabular}{lccccc}
		\toprule
		& $\;$doors$\;$ & $\;$NYU2$\;$ & Pasadena & Omni\textsc{Inst} & Omni\textsc{Symb}\\
		\midrule
		Baseline                                     & 86.1 & 76.7 & 85.0 & 84.4 & 72.3\\
		LtC,$\,k$=10                            & 64.3 & 72.9  & 79.7 &81.5&66.9\\
		\midrule
		Siamese,$\,$avg,$\,k$=1               & 90.6 & 82.2 & 90.7 & 94.6 & 79.6\\ 
		Siamese,$\,$avg,$\,k$=10             & 90.9 & 83.3 & 90.7 & 94.6 & 79.8\\
		Siamese,$\,$avg,$\,$all           & 91.1  & 82.2 & 89.5\tablefootnote{Using all ($>$2000) examples not tractable, we set $k$=100.} & 94.3 & 79.9\\
		\midrule
		Siamese,$\,$vote,$\,k$=10             & 91.0   & 83.3  & 89.8 & 94.7 &79.9 \\
		Siamese,$\,$vote,$\,$all  & 90.7   & 82.2 & 89.5\footnotemark[3] & 94.7 &	79.8 \\ 
		\bottomrule
	\end{tabular}
	\caption{Classification accuracies on different datasets.}
	\label{table1}
\end{table}

%

\vspace{0.5em}
\noindent
\textbf{Open and Closed Doors} is a new dataset collected for the
experiment that sparked this work and served as running example in the
paper. The goal is to determine whether a door is \emph{closed} or
\emph{open}. The dataset consists of RGB-D images acquired by a real
mobile robot (Fig.~\ref{fig:0}), and includes hinged doors and
roll-up doors.
There are multiple images per door with varying lighting and
open/closed status conditions, recorded at two different locations
(physically different warehouses) and with an unbalanced class
distribution with only $\approx$25\% open instances (in both
locations).
We designate one location as training set, and the other one as test
set. Since each location only has a limited number of doors (11 doors
/ 850 images for training, respectively 10 doors / 792 images for
testing) which vary in appearance between locations, there is a clear
domain shift and we expect conventional classification to overfit to
the closed world of the training location.

As can be seen from the results in Table \ref{table1}, similarity
learning significantly increases the classification performance,
reducing the mis-classification rate from 13.9\% to $\approx$9\%.
In more detail, the baseline and the Siamese similarity network give
the same answer for 74.0\% of the test set. But in 18.3\% of cases the
Siamese network is right when the baseline is wrong, whereas the
opposite is true only for 7.0\% of the data. See Fig.~\ref{fig:2.5}
for examples.
Visually, one can see that the Siamese network handles bad lighting
conditions better, whereas the direct classification apparently relies
too strongly on the RGB image and fails if it is under-exposed, which
happens sometimes at the test location. Conversely, the Siamese
network is seemingly more confused by strong artifacts in the depth
channel.
Surprisingly, both methods are wrong only on 0.6\% of
the data, which suggests that they are to a large degree
complementary. It is a promising future direction to explore whether
this can be exploited by somehow combining them.

Using more random exemplars per class only slightly improves
performance. Even at $k\!=\!1$ the method performs almost as well as
when considering similarities to all training examples, indicating
that the CNN embedding separates the two classes rather well.

\makeatletter
\newlength{\failwidth}
\setlength\failwidth{0.32\columnwidth}
\makeatother
\begin{figure}[t]
	\centering
	\begin{tabular}{ccc}
		\includegraphics[width=\failwidth]{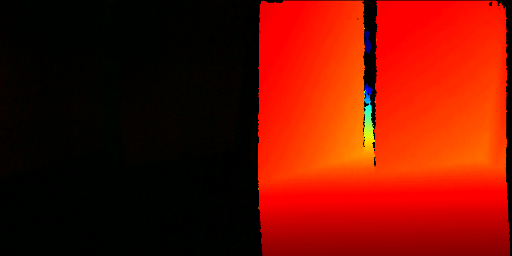} &
		\includegraphics[width=\failwidth]{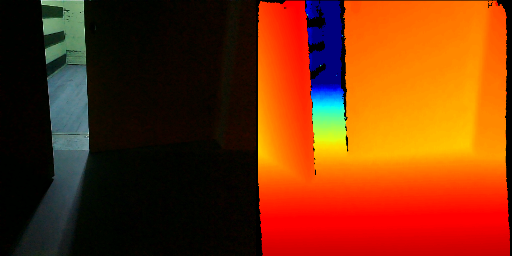} &
		\includegraphics[width=\failwidth]{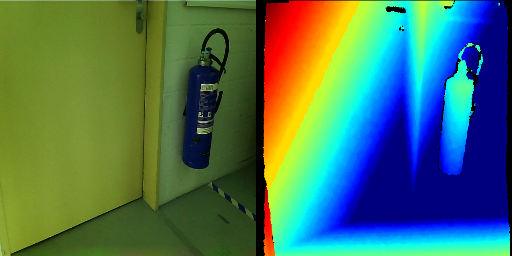}\\
		\raisebox{0.5em}{Siamese correct} & \raisebox{0.5em}{Siamese correct} & \raisebox{0.5em}{Siamese wrong}\\
	\end{tabular}
	\vspace{-0.7em}
	\caption{Examples cases where direct classification and
		similarity learning predict different classes.}
	\label{fig:2.5}
\end{figure}

\makeatletter
\newlength{\nyuwidth}
\setlength\nyuwidth{0.48\columnwidth}
\makeatother
\begin{figure}[t]
	\centering
	\begin{tabular}{ccc}
		\includegraphics[width=\nyuwidth]{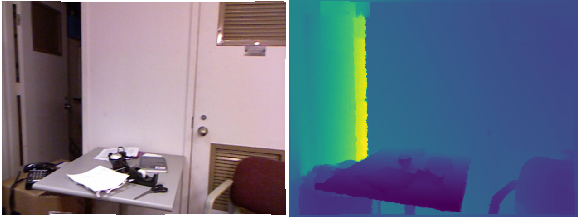} & &
		\includegraphics[width=\nyuwidth]{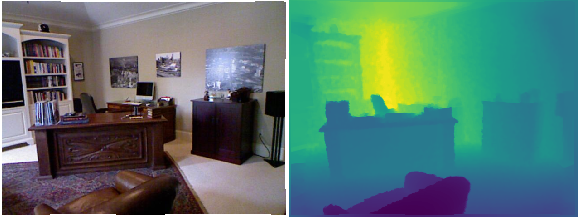}\\
		\raisebox{0.5em}{office} & & \raisebox{0.5em}{home office}\\
	\end{tabular}
	\vspace{-0.7em}
	\caption{Example images from NYU2 RGB-D dataset.}
	\label{fig:3}
\end{figure}
%
\vspace{0.5em}
\noindent
\textbf{NYU2} is a public dataset \cite{Silberman:ECCV12} that
consists of RGB-D images showing different types of indoor scenes
(like ``kitchen'', ``hallway'', etc.). We select the rather
challenging task to tell apart the two classes \emph{office} and
\emph{home office}. Since both are GPS-denied indoor environments,
this may for instance be useful for self-localisation.
There are 50 instances of home office and 78 instances of office in
the dataset. We train on 30\% of them (16 home office, 26 office) and
then test on the remaining ones.
Since the dataset is very small, we randomly split into train and test
portions and trust that there will be some degree of domain shift
between them, where the training set exhibits unintended biases not
replicated in the test data.
Due to the small amount of data, we reduce the minibatch size to 8.

Also for this dataset, the Siamese approach reduces the error rate,
from 23.3\% to $\approx$16.5\%. Interestingly, a moderate number
$k\!=\!10$ of exemplars seems to be working slightly better than using
fewer, or more. This remains to be investigated further in future
work.

\makeatletter
\newlength{\treewidth}
\setlength\treewidth{0.27\columnwidth}
\makeatother
\begin{figure}
	\centering
	\begin{tabular}{cccc}
		\includegraphics[width=\treewidth]{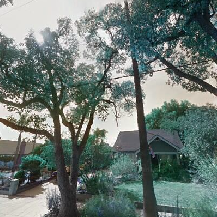} &
		$\quad$ &
		\includegraphics[width=\treewidth]{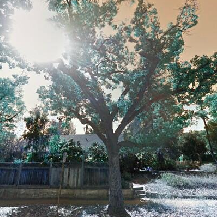}\\
		\raisebox{0.5em}{Mexican Fan Palm} & & \raisebox{0.5em}{Camphor Tree}\\
	\end{tabular}
	\vspace{-0.7em}
	\caption{Example images from Pasadena street trees
		dataset.}
	\label{fig:5}
\end{figure}
%

\vspace{0.5em}
\noindent
\textbf{Pasadena Street Trees} \cite{wegner2016cataloging} provides
RGB images of different tree species, cropped from Google Streetview
panoramas. We define an environment mapping task, to distinguish the
two frequent species \emph{Camphor Tree} (6745 examples) and
\emph{Mexican Fan Palm} (7595 examples). We train on 30\% of both
classes and then test on the data rest, chosen at random.

This dataset is a good example that even training sets of apparently
reasonable size may not be enough to ensure a stable domain, especially
when operating in the wild under weakly controlled conditions.
Due to the high variability of both the trees and the background, the
Siamese similarity learning significantly outperforms a direct class
prediction, even though there are $>$2500 training images per class.
The error rate is reduced from 15\% to $\approx$9.5\%. Due to the
comparatively big volume of data, it is not tractable to compute the
similarities to all training examples, hence we approximated them by
setting $k\!=\!100$, which still takes $>$3 seconds per test image on a
Titan Xp GPU.
Fortunately, the experiment again confirms that small exemplar sets
$k\!<\!10$ are sufficient to achieve good results.

\makeatletter
\newlength{\omniwidth}
\setlength\omniwidth{0.48\columnwidth}
\makeatother
\begin{figure}[t]
	\centering
	\begin{tabular}{ccc}
		\includegraphics[width=\omniwidth]{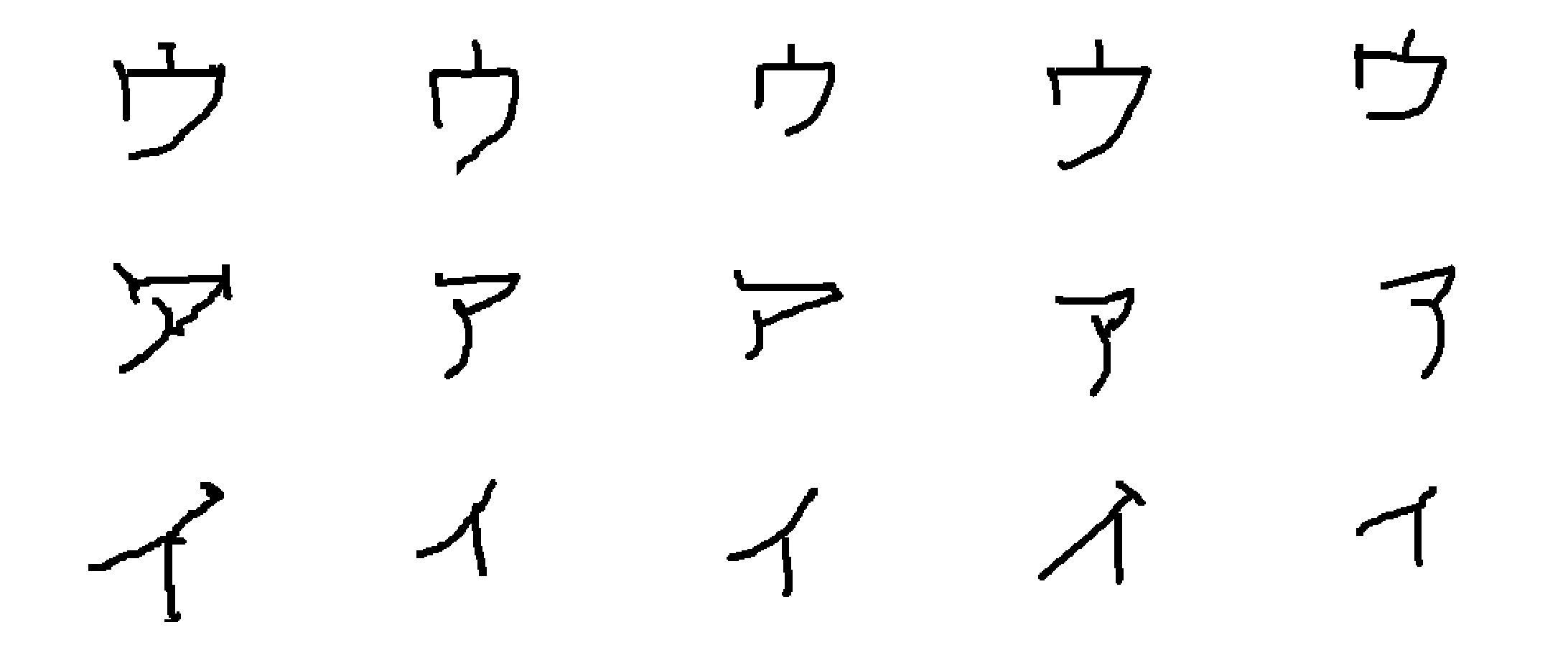} & &
		\includegraphics[width=\omniwidth]{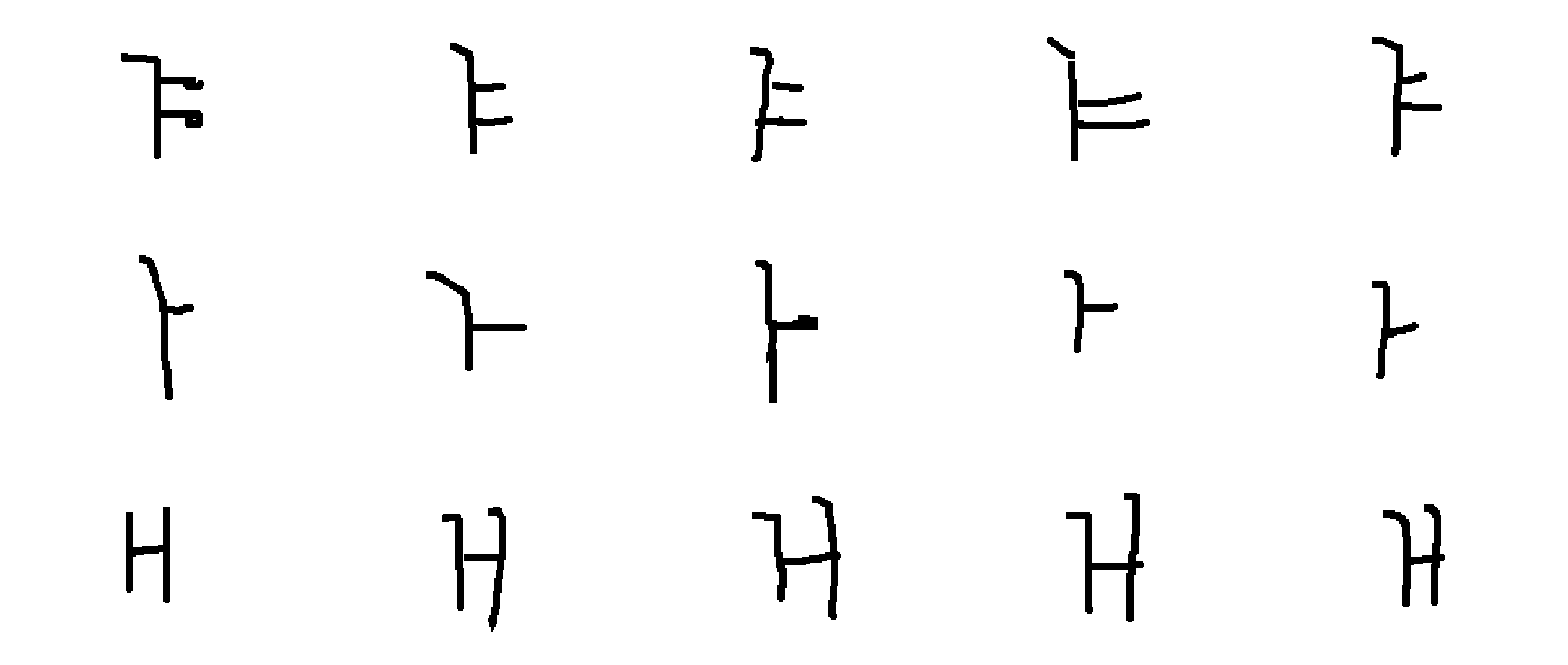} \\
		\raisebox{0.7em}{Katakana} & & \raisebox{0.7em}{Korean}\\
	\end{tabular}
	\vspace{-0.7em}
	\caption{Examples of Omniglot characters. Each row shows multiple
		instances of the same character. The two test set-ups correspond
		to sampling only some columns (\textsc{Inst}) or only some rows
		(\textsc{Symb}) for training.}
	\label{fig:6}
	\vspace{-0.5em}
\end{figure}
\makeatletter
\newlength{\sigwidth}
\setlength\sigwidth{0.45\columnwidth}
\makeatother
\begin{figure}[t]
	\centering
	\begin{tabular}{cc}
		\includegraphics[width=\sigwidth]{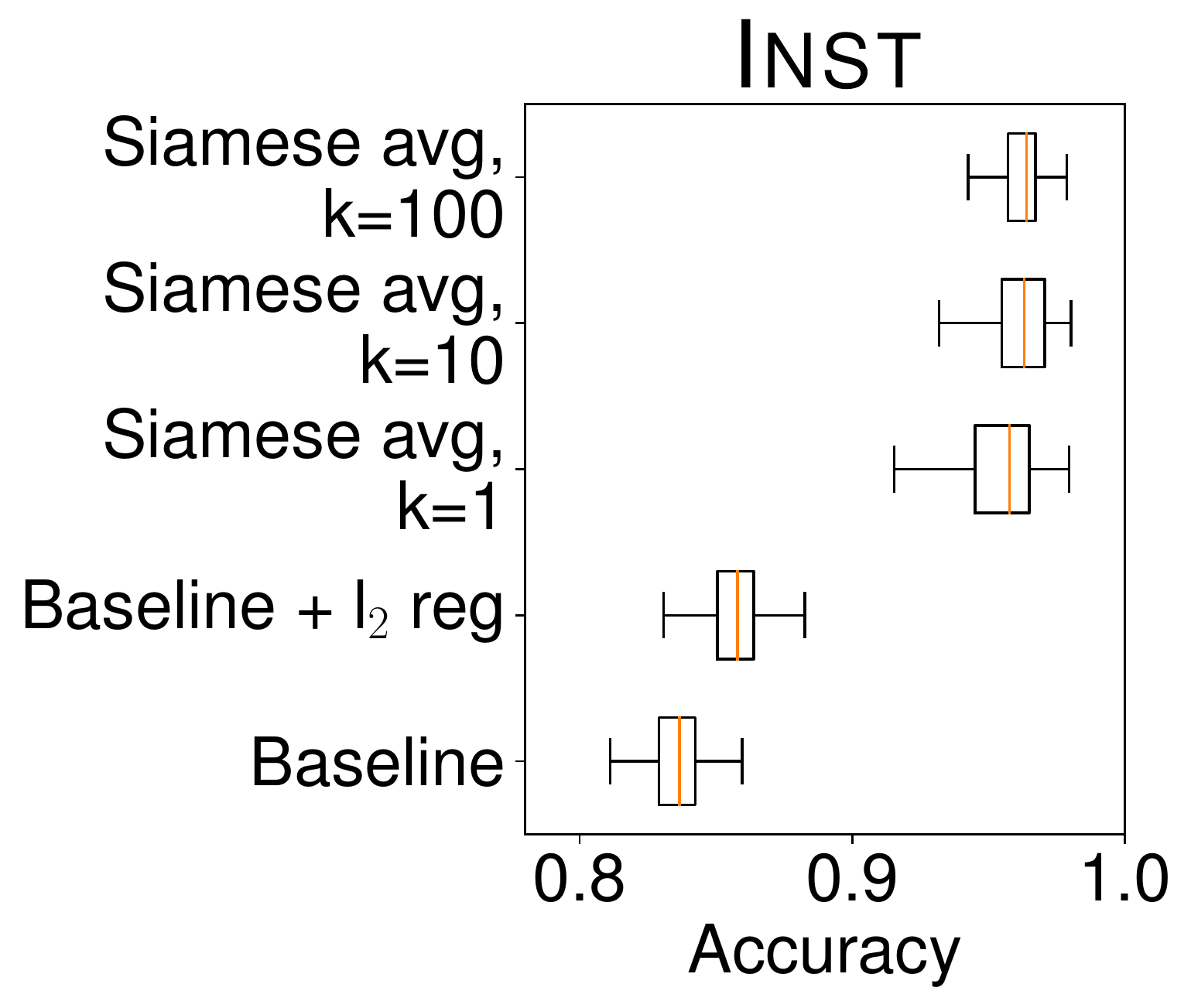}&
		\includegraphics[width=\sigwidth]{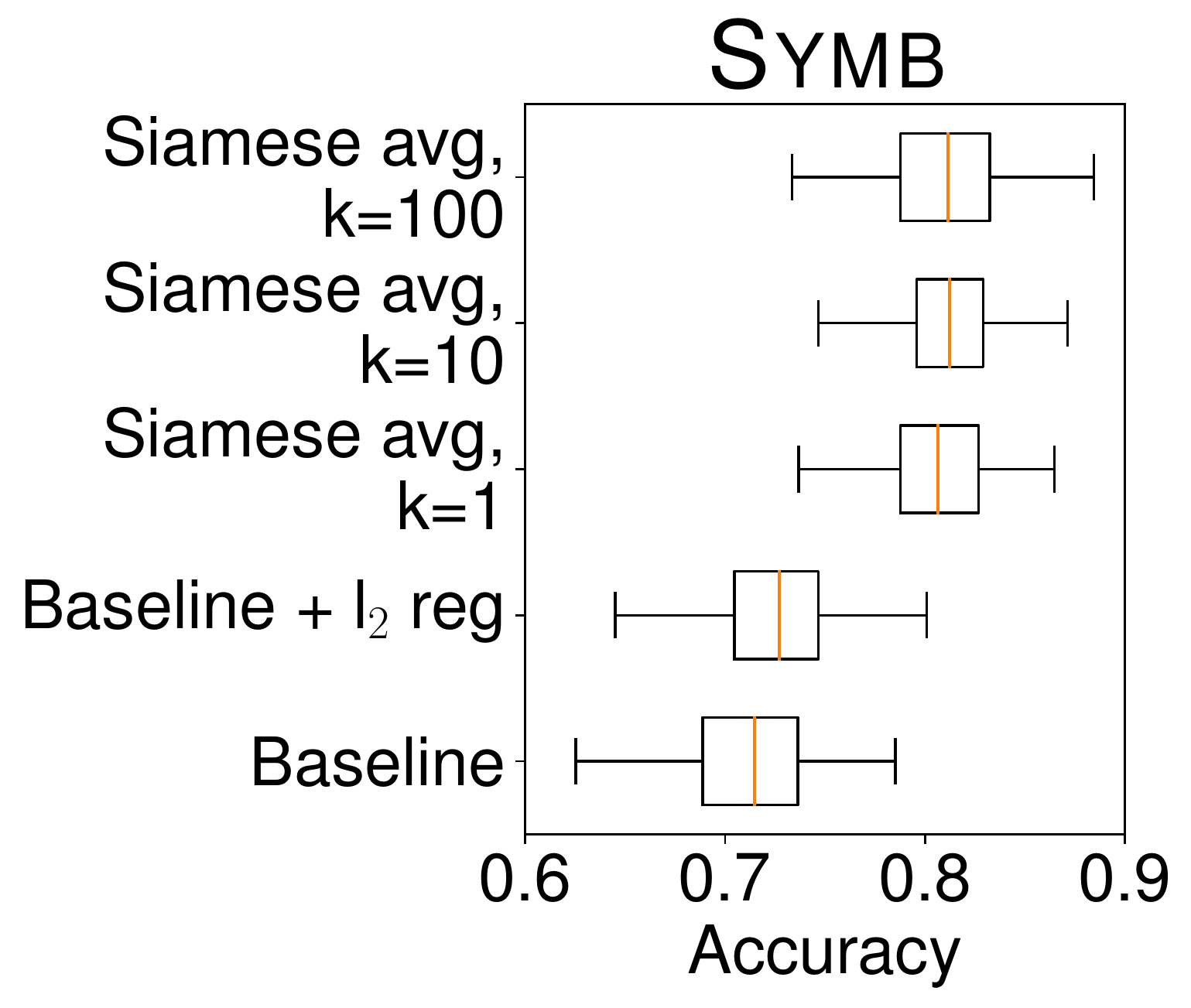}
	\end{tabular}
	\vspace{-0.5em}
	\caption{Stability under varying train/test splits, and comparison to
		regularised, direct classification.}
	\label{fig:6.1}
\end{figure}
%
%
\vspace{0.5em}
\noindent
\textbf{Omniglot} \cite{lake2015human} is one of the most popular
datasets for distance, respectively similarity learning. It consists
of greyscale images of handwritten characters from 50 different
alphabets. For each character there are 20 different instances
(writers).
We define two different tasks with domain shift. In both cases the
goal is to distinguish Korean characters from Japanese Katakana
characters. The alphabets were chosen because they are known to be
visually similar and therefore hard to distinguish. The two tasks are
defined as follows:
\begin{itemize}
	\item\textsc{Inst}: randomly select 280 Katakana + 240 Korean
	instances (30\% of total) for training, and the remaining 660 +
	560 for testing. I.e., the domain shift arises from the fact that
	7 writers are not enough to capture all legitimate variations in
	writing style. Exemplars are drawn at random, so the chance is low
	($<4$\%) that a test image is paired with a training image of the
	same character. Still, the network has seen all characters of both
	languages during training.\vspace{0.3em}
	\item \textsc{Symb}: randomly select 14 Katakana + 12 Korean
	characters (30\% of total) for training and the other 33 + 28 for
	testing. This creates a more difficult domain shift: the
	network must learn the ``stylistic commonalities'' of Japanese,
	respectively Korean characters from only 30\% of the alphabet,
	such that they generalise to the remaining 70\% of
	which the network has never seen any instance.
\end{itemize}

On this dataset the Siamese approach does particularly well. For the
\textsc{Inst} scenario it reduces the error rate by two thirds, from
15.6\% to $\approx$5.5\%.
For the particularly difficult \textsc{Symb} scenario, the error rate
drops from 27\% to $\approx$18.5\%.
Recall that the latter is indeed a very challenging problem: the
classifier only ever sees 30\% of the symbols from the two alphabets,
and must learn to assign a symbol it has never seen before to the
right alphabet. Under these circumstances, the performance is quite
remarkable, compared to chance level.

\begin{table}[b]
	\centering
	\begin{tabular}{lccccc}
		\toprule
		&  Baseline & avg & voting & LtC  \\
		\midrule
		k=1               & 84.4 & \textbf{94.6} & --- & 77.9 \\ 
		k=10             & 84.4 & 94.6 & \textbf{94.7} & 81.5 \\
		k=100 (all for our method)  & 84.4 & 94.3 & \textbf{94.7} & 80.7\\
		\bottomrule
	\end{tabular}
	\caption{Comparison to few-shot learning on Omni\textsc{Inst}.}
	\label{table2}
\end{table}

We have argued that heavy regularisation of the direct classification
baseline cannot be expected to overcome domain shift. To test this
claim, we ran the baseline without regularisation and with
$l_2$-regularisation of the weights. The strength of the regularizer
is tuned for best performance by grid search.
To exclude biases due to a particular train/test split and to assess
how significant the differences are, we perform 100 different random
splits, for both the \textsc{symb} and \textsc{inst} task.
In Fig.~\ref{fig:6.1} one can see that regularisation only slightly
improves the classification at test time, and that the advantage of
the Siamese similarity approach is persistent across different splits.
This supports our assertion that regularisation is not enough
to cope with domain shift.

In few-shot learning terminology, our method is a form of the
\emph{2-way, k-shot} scenario. As further baseline, we therefore run
the ``Learning to Compare'' (LtC) \cite{zagoruyko2015learning} code.
We tested three settings for LtC on Omniglot\textsc{Inst},
$k\in[1,10,100]$. We find that LtC cannot beat the direct baseline
even when choosing a high $k\!=\!100$, and never reaches the
performance of our simple Siamese network. See
Tab.~\ref{table2}. Experiments on other datasets confirm this
result, see Tab.~\ref{table1}.

The experiment illustrates the difference between the domain shift and
few-shot problems. In our setting, we do have a larger training set of
class examples, which are all used to learn the best possible
similarity function. Only at test time we sample a small set $k$, for
efficiency and empirically without loss of accuracy.
Whereas LtC learns a similarity mostly from other classes, which is
then tuned to only $k$ exemplars of the target classes. Therefore, it
cannot recover the data manifold as well, and has less knowledge of
the intra- and inter-class variations beyond the exemplars.
Moreover, LtC consumes more GPU memory, because the complete support
set is stored for the forward pass.


\section{Conclusion}
We have argued that a machine vision system that operates in the wild
will in some situations face visual domain shift, since it is not
always possible to capture all the variability of the system's future
environment at training time.
We have investigated similarity learning with a Siamese CNN as a way
of learning classifiers that perform well in the presence of domain
shifts. We found that the network embeds the training data in such a
way that they are well separated and (relative) similarities are
fairly reliable between \emph{arbitrary} pairs of data points. Hence,
an unseen test case can be classified by sampling similarities to few
random exemplars from the training data. In experiments on four
datasets, deep similarity learning consistently outperforms direct
classification.

There are several open points that merit further investigation. First,
our study was limited to binary classification. While it is
conceptually straight-forward to generalise the idea of similarity
learning to multiple classes, it is much less clear how to best
implement it.
One possibility is to decompose it into multiple pairwise
similarities, which will however exponentially increase the number of
pairings that must be trained.
Another idea would be to directly learn a set of similarities, or a
ranking, using exemplars from all classes as additional input.
A second point concerns the use of multiple exemplars. Although in our
study sampling $k\!>\!1$ exemplars brought only minor improvements, it may
be a mechanism to robustify the classification in scenarios where the
classes are not well separable.
In that situation the question arises how to best combine the
per-exemplar similarities. Here, we have tested straight-forward,
handcrafted averaging and voting schemes. It may however be
interesting to also learn the combination, or even to explore an
``early combination'' where a multi-way
similarity~\cite{hartmann17iccv} is computed from a test example to a
set of multiple exemplars.

\bibliographystyle{unsrt}
\bibliography{../literature}

\end{document}